\documentclass[conference]{IEEEtran}
\IEEEoverridecommandlockouts
\usepackage{cite}
\usepackage{amsmath,amssymb,amsfonts}
\usepackage{algorithmic}
\usepackage{graphicx}
\usepackage{textcomp}
\usepackage{xcolor}
\def\BibTeX{{\rm B\kern-.05em{\sc i\kern-.025em b}\kern-.08em
    T\kern-.1667em\lower.7ex\hbox{E}\kern-.125emX}}
\begin{document}

\title{ Using Keypoint Matching and Interactive Self Attention Network to verify Retail POSMs\\
}

\author{\IEEEauthorblockN{Harshita Seth}
\IEEEauthorblockA{
\textit{ParallelDots, Inc.}\\
harshita@paralleldots.com }
\and
\IEEEauthorblockN{Sonaal Kant}
\IEEEauthorblockA{
\textit{ParallelDots, Inc.}\\
sonaal@paralleldots.com}
\and
\IEEEauthorblockN{Muktabh Mayank Srivastava}
\IEEEauthorblockA{
\textit{ParallelDots, Inc.}\\
muktabh@paralleldots.com }
}


\maketitle

\begin{abstract}
Point of Sale Materials(POSM) are the merchandising and decoration items that are used by companies to communicate product information and offers in retail stores. POSMs are part of companies' retail marketing strategy and are often applied as stylized window displays around retail shelves. In this work, we apply computer vision techniques to the task of verification of POSMs in supermarkets by telling if all desired components of window display are present in a shelf image. We use Convolutional Neural Network based unsupervised keypoint matching as a baseline to verify POSM components and propose a supervised Neural Network based method to enhance the accuracy of baseline by a large margin. We also show that the supervised pipeline is not restricted to the POSM material it is trained on and can generalize. We train and evaluate our model on a private dataset composed of retail shelf images. 

\end{abstract}

\section{Introduction}
Computer Vision is used in many business applications nowadays especially Fast Moving Consumer Goods (FMCG) companies are using computer vision for detecting and recognizing products in supermarkets. This helps them evaluate their retail presence with respect to their competitors. Computer Vision can also have applications in retail marketing and merchandising. Point of Sale Material (POSM) are advertising materials that are used to communicate product information and discounts to the consumers in retail stores. FMCG companies want to be assured of the fact that all merchandising material is being placed according to their specifications so that their consumers can be aware of the latest offers and new products. In this work, we have proposed Computer Vision methods which can be used to automatically verify POSMs from retail shelf images. 

Most POSMs are applied as window displays that is stylized windows and shelves in supermarkets. These grab customer attention as they are searching for products on shelves. Each window display is supposed to have many components like cutouts and shelf strips. We aim to detect whether a photograph belongs to a specific POSM as well as verify the presence of all components of window displays in the image.

\begin{figure}[t]
    \centering
    \includegraphics[width=200pt]{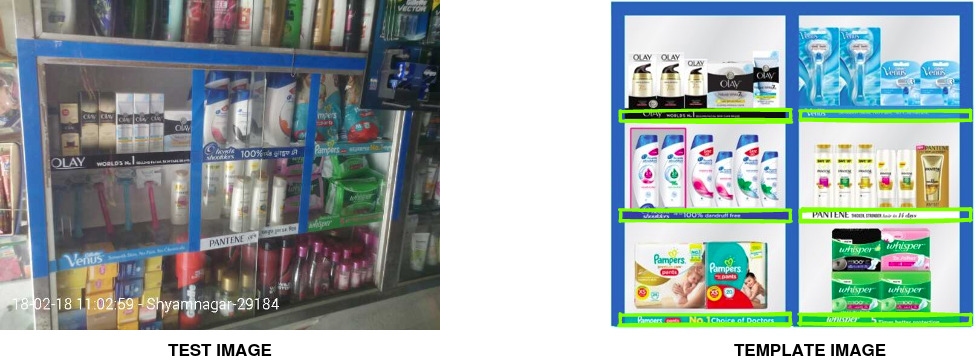}
    \caption{\textbf{Left:} A real world test image from inside of a retail store, \textbf{Right:} An template image for a window display POSM. Our Aim is to detect the complete POSM in the real world image along with the individual presence of shelf strips [and other applicable parts] which are shown using green bounding boxes. Such bounding boxes are available for all parts in all template images. Readers can notice the noise and perceptual difference between templates and real world photos of POSMs}
    \label{fig:demo}
\end{figure}

A Window display is a combination of shelf strips and cut-outs which we refer to as parts in this paper. As shown in fig \ref{fig:demo} a Window display in a canonical domain is referred to as template image. The template is computer Generated using software like inkscape or illustrator. On the other hand the test image is a real world photo taken from inside of a shop. The discrepancy between the real domain and canonical domain can be seen in fig \ref{fig:demo}. Apart from computer generated template looking different from a real world instance, the relative dimensions of different parts can also vary due to variable sizes of retail shelves across stores, making the verification task non-trivial. To deal with the large perceptual gap in the visual domain we use a keypoint matching based approach that works on local feature matching. we use this to find a POSMs of template image in a real world test image. We divide our task into two parts : 1. Detecting whether a POSM is present in an image and 2. Verifying whether all parts of a POSM are individually present. For POSM detection, we use CNN based keypoint matching. In our baseline for verifying individual parts, we show that simple rules on CNN keypoint matching methods can give us a good initial performance. As an improved method of individual part verification, we replace simple rules on keypoint matching by a neural network called Interaction Network that enhances the accuracy. Given a template image (T) and real world test image(q), keypoint matching is performed on both to retrieve the matched local features which are then used by Interaction Network for classification of parts as found or not found. The key ideas when attempting to alleviate the above challenges of part verification and domain discrepancy are 1) We use Superpoint\cite{b1}, a local feature based keypoint matching technique to identify POSM of the template in real world test image. 2) We use attention based method for template part presence verification. 

\begin{figure*}[ht]
    \centering
    \includegraphics[width=400pt]{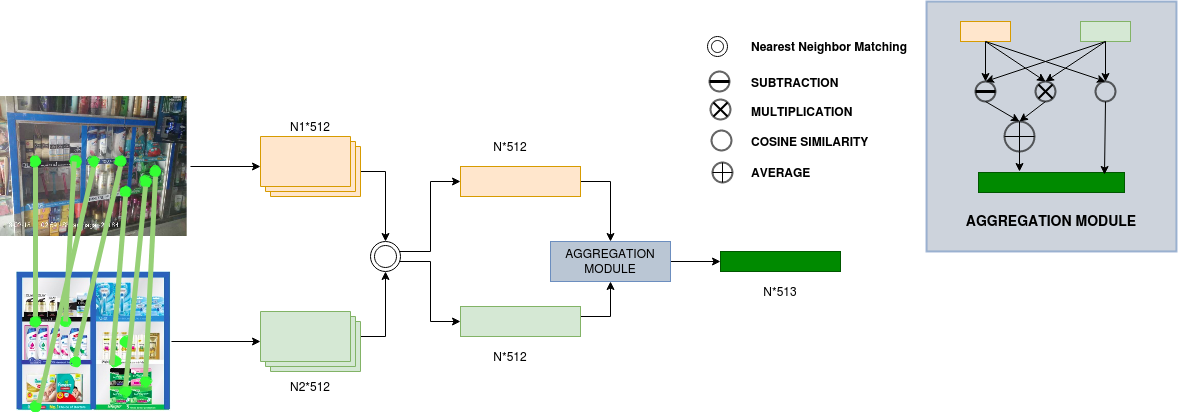}
    \caption{Overview of our first step : N1 and N2 are the descriptors and keypoints from the superpoint architectures. After nearest neighbor filtering, N matched pairs are then passed through the aggregation module as shown in blue box. The final list of embeddings of template and real world test image is then further used in second step of interaction network.}
    \label{fig:full_arch}
\end{figure*}

\section{{Related Work}}

Template matching is one of the most frequently used techniques in computer vision  and is very closely related to our problem as given a template image we have to locate it in a real world test image but all the conventional template matching techniques fail due to two reasons 1.Our problem is not limited to just matching, we also aim to detect the presence of each part of POSM in real world test image, 2. The relative dimensions of POSM parts in real world stores is not necessarily same as in template. 

We divide our problem into two subproblems 1) Global Matching and 2) Local Part Matching.

\textbf{Global Matching :} Template matching is a classical problem which was solved using basic machine learning methods like sum-of-squared-differences or normalized cross correlation to calculate the similarity score between the template and the real world test image. But these methods were limited to very small transformation between template and real world test image which was later improved by Dekel et al.[11] that introduced the measure to remove the bad matches caused by background pixels. As an improvement DDIS \cite{b11} was introduced by Talmi et. al, that uses multiple template deformation before nearest neighbor matching. These classical methods fail to perform when there is complex transformation or huge domain variance.
Later many deep learning based approaches\cite{b3},\cite{b5},\cite{b7},\cite{b6},\cite{b4} were introduced for stereo matching, object tracking etc. that uses shared deep architectures for feature extraction and performs feature matching on template and real world test image to get a similarity score. 
All the recent state of the art template matching algorithms uses deep features to locate the complete template image in the test image and none of the above methods can be used for part level presence search. 

\textbf{Local Matching :} Local feature matching is a vast field of research which aims to recognize features of the same object across different viewpoints and domains. The preliminary step of local feature matching is detecting the interest points, referred as KeyPoints. Traditional Interest point detectors such as Harris Corner Detection \cite{}, FAST \cite{b12} are very well known. As a second step for local matching a descriptor for each interest point is created which is an information that stands apart from other keypoints. Traditional and famous algorithms such SIFT \cite{b9}, ORB \cite{b10} and many more are used for local descriptors but recently many deep learning approaches have been introduced which outperforms the traditional machine learning algorithms for keypoint detection and descriptor matching. De tone et.al \cite{b1} introduced a multi task deep learning algorithm for both the keypoints and descriptors known as Superpoint. We use this technique as a baseline for our problem and introduce the certain challenges which are faced by this algorithm such as high dependency on threshold for number of matches, huge domain shift in template and test image. We try to overcome these challenges by introducing another network, called Interaction Network explained in detail in section \ref{fig:interaction_network}, on top of superpoint.

\textbf{Self-Attention:} Our interaction network is inspired by self attention \cite{b13}. A self-attention module responds to a position of a sequence by taking into account the information of all positions and taking their weighted average in an embedding space. We have used to capture the interaction between the part descriptors. Self-attention is predominantly used in machine translation \cite{b15}\cite{b16}\cite{b17} but have also been extended to image and video problem in computer vision. Non-local operation \cite{b14}, scene segmentation \cite{b18}, classification \cite{b19} are one of them.

\section{OUR APPROACH}
\label{fig:our_approach}
As of our knowledge this is the first time anyone is trying to solve this real world problem statement. Recently \cite{b2} has introduced a slightly similar problem statement of classifying Leaflets promotion but those scenarios are comparatively easy to solve as they are digital and we work on real world images taken from a smartphone.  
We aim to solve this problem without any requirement on training on new Window displays. We use our in-house dataset to prove the following : 1. Fully unsupervised CNN keypoint matching can act as a good baseline and 2. interaction network trained on in-house data can generalize to other unseen templates. In our training dataset, we have a real world test image which is mapped to a template image. Fig \ref{fig:demo} shows the  pair of image, test(a)  and template(b).  Ground truth for the pair of images is the presence of each part in the real world image and bounding box annotation of the corresponding part in the template image. Our training data consist of 300k real world test images which is mapped to 37 unique gallery images, we have used 250k images for training and remaining as validation data. To
test model generalization we have created separated test data which have 21k real world test images which are mapped with 8 template images. The intersection between train template image and test template image is zero.


Our approach presented in this paper is focused on detecting the presence of template image and its parts as shown in the fig \ref{fig:demo}. We divide the solution into the following two main parts:
\begin{enumerate}
    \item Detection of POSM as a whole in real world test image image using keypoint based methods
    \item Verification of the parts of the template image using simple rules on keypoint matching in baseline and training a interaction network on output of keypoint matching as an enhancement.
 
\end{enumerate}
\begin{table}
\begin{center}
\caption{POSM Detection}
\begin{tabular}{ |c|c|c|c|c| } 
\hline

Method & Accuracy & F1 & Recall & Precision \\
 \hline
 Superpoint   & 0.747  & 0.838 & 0.967 & 0.739  \\
\hline

\end{tabular}
\label{table:Posm_detection}
\end{center}

\end{table}
\subsection{Detection of POSM}
POSM is detected by using CNN keypoint matching. This verifies whether an real world image contains a POSM. The proposed keypoint detector and descriptor method is based on De tone et. al. \cite{b1} as it has been proven to outperform SIFT\cite{b9}, ORB\cite{b10} based methods significantly. Superpoint has been the state of the art on many datasets and is known best for local feature matching. We use Superpoint as our base Keypoint detector and descriptor. Given a template image (T) and test image(q) we pass both the images through the pretrained superpoint model to extract all the relevant local features. 
The above architecture is used to detect descriptors N1 and N2 of the test and template image respectively. N pairs of descriptors are extracted using an exhaustive nearest neighbour search algorithm. The unmatched points and descriptors are discarded from both the images. We decipher simple rules about N to determine the presence of POSM in real world images. If the number of matched points N between template and real world image is greater than a threshold t , we say that the POSM is present in the real world image. While determining the threshold t from a sample of images, we keep into mind the possibility of POSM parts missing from the real world images.

\subsection{POSM part Detection}
After matching keypoints of a template and a real world image, we get pairs of matched points which like stated in above section are used to check for presence of POSM. Next we define the problem of detecting individual POSM parts and how the matched keypoints are used to perform the task.
For a template $n$ \& real world image pair, data for $i$th pair of matched keypoints can be represented as $ \{K^t_{ni},K^r_i,D^t_{ni},D^r_i\} \in [K^T,K^R]$, where $K^t_n \text{ and } K^r_i \in R^2_i$, a 2D image coordinate comes from template image and real world image respectively, and similarly $D^t_{ni} \text{ and } D^r_i \in R^n$ are their descriptors. Each POSM template can have multiple parts like shelf strips, cutouts, posters etc. In our dataset, we have annotations for bounding boxes of all parts of POSMs in their template images \ref{fig:demo} . Let the $m$th part of nth POSM be denoted by $P^m_n$ and its bounding box is $B^m_n \in R^4$. Thus, each $K^t_{ni}$ can be determined to be present in the part $ \text{(whether } K^t_{ni} \text{from }P^m_n \text{)} \text{ if } K^t_{ni} \in B^m_n$

When training a supervised learning algorithm, we collate all the keypoints pairs and their descriptors to their respective parts. We give two labels to each matched pair 1)  Part  in whose bounding box in template image keypoint from a matched pair lies  or "None" if the keypoint doesn't belong to any part. 2) Another global label we give is 1 if POSM is present else the global label given is 0. So we now have a cartesian of the form
{$K^t_{ni},K^r_i,D^t_{ni},D^r_i, P^m_n=0/1, POSM \text{ present }=0/1$}.

We use these labels to calculate the loss of our interaction network.
\subsubsection{POSM part verification using simple rules on matched keypoints}
\label{rules}
We propose some simple rules which we can apply on the N matched keypoint pairs of template-test image to verify parts of POSM. In our dataset we have ground truth of the presence of each part in the real image and the bounding box annotation of the corresponding part in the template image. Using those annotations we count the number of keypoints that got matched and also lie inside the annotated bounding box for each part of the template image. If the count of those keypoints inside a part is greater than the threshold $t_p$ then we predict that part to be present in the real image. We use this simple rule for part verification as our baseline. So if for a template/real world image pair
n, $ \{K^t_{ni},K^r_i,D^t_{ni},D^r_i\} \in [K^T,K^R]$ where $K^t_{ni}$ of part $P^m_{n}$ > $t_p$
that part is deemed to present in the real world image.

\begin{figure}[t]
    \centering
    \includegraphics[width=250pt]{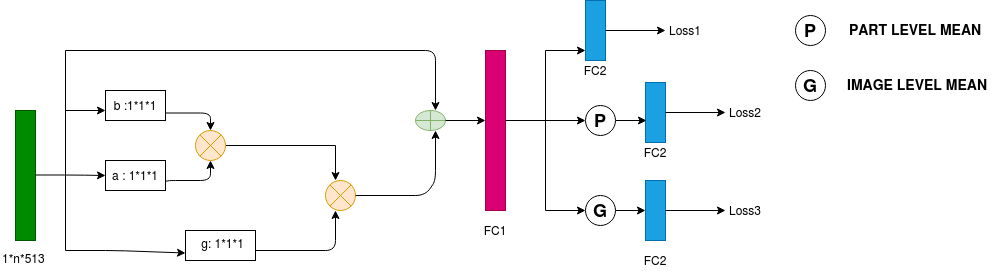}
    \caption{Overview of Second step : A self attention module is applied on the output of the first step followed by couple of linear layers. There are three loss in this step which are applied on shared fc2 layer. }
    \label{fig:non_local}
\end{figure}

\begin{table*}[t]
\centering
\caption{}
\begin{tabular}{llllllllll}
\hline
\multicolumn{3}{c}{Loss}                                                          & \multicolumn{3}{c}{Aggregation}                                                   & Rec   & Prec  & Acc   & F1    \\ \hline
Loss1                     & Loss2                     & Loss3                     & Similarity                & Multi                     & Concat                    &       &       &       &       \\ \hline
\checkmark & \checkmark & \checkmark & \checkmark & \checkmark &                           & 0.778 & 0.956 & 0.832 & 0.858 \\ \hline
\checkmark & \checkmark & \checkmark &                           & \checkmark &                           & 0.794 & 0.889 & 0.801 & 0.839 \\ \hline
\checkmark & \checkmark & \checkmark &                           &                           & \checkmark & 0.684 & 0.958 & 0.774 & 0.798 \\ \hline
\checkmark & \checkmark &                           & \checkmark & \checkmark &                           & 0.767 & 0.952 & 0.831 & 0.856 \\ \hline
                          & \checkmark & \checkmark & \checkmark & \checkmark &                           & 0.774 & 0.959 & 0.831 & 0.857 \\ \hline
                          & \checkmark &                           & \checkmark & \checkmark &                           & 0.772 & 0.960 & 0.830 & 0.856 \\ \hline
\end{tabular}
\label{table:ablation}
\end{table*}

\subsubsection{Interaction Network based POSM part verification}
\label{interaction_network}

We propose an Interaction Network which is used for two main purpose 1) To detect parts of a POSM as an alternative to simple rules we used in the baseline 2) To alleviate the discrepancy in canonical domain of template image and real world domain of real image using an attention based mechanism.
\subsubsection{Self Attention}
 Attention was first introduced by \cite{b13} to help memorize the long sentences in Neural Machine translation. The output of the attention module is the weighted sum of the value where each weight is determined by the softmax on the dot product of Query and Key.\\
 
$ Attention(Q,K,V) = softmax(\frac{QK^T}{\sqrt{d_{k}}})V$ \\
 
 Self Attention is when both Query, Key and Value come from the same input. We use self attention to learn combinatorial inter dependencies among representations of all pairs of matched descriptors. Please note that while self attention is generally used on numbered sequences where each item of sequence is enumerated and is given a positional embedding, modelling matched keypoints is unordered and thus we don't use any positional embeddings. 


The output from the superpoint network gives N pairs of matched descriptors for each pair of template/real world images $[K^T,K^R]$. Instead of passing the pair of these features directly to the interaction network we perform some aggregating operations on their decriptors and pass them as input. As shown in figure \ref{fig:full_arch}, for a given a pair of descriptors we create a single embedding of length 513 using subtraction, multiplication and similarity operations on their decriptor pair. After concating each 513 descriptors $D^{t^r_{i}}$, [$D^{t^r_{1}} \| D^{t^r_{2}}\|............ D^{t^r_{i}} \|$........$D^{t^r_{N-1}} \| D^{t^r_{N}}$] we have list of N embeddings.

The final list of N embeddings of length 513 are passed through the self attention module in which three 1d convolutions are used for deriving Query, Key and Value from the embeddings. The attention module computes responses based on relationships between different locations which are not captured in a fully connected layer. After the attention module we use a couple of fully connected modules FC1, FC2 for classification of each embedding.

We use the labels we have assigned above for each descriptors for the loss calculation. We introduce three cross entropy loss function for the proposed interaction network.1) Part Loss ($P_{loss}$) :  Out of the two assigned labels, the first one is used for part classification. The mean of all the embeddings that belong to a particular part is pass to a shared fully connected layer (FC2) for classification. 2) POSM loss ($G_{loss}$) : The POSM loss is applied to classify the presence of complete POSM using the global mean of all embeddings. 3) Embedding loss ($E_{loss}$) : Additional embedding classification loss is added to make the network learn the difference between the wrong matched pairs ( False Positives) and the correct matched pairs ( True Positives).




\begin{table}
\begin{center}
\caption{Taking unmatched part count as zero}
\begin{tabular}{ |c|c|c|c|c| } 
\hline
Method & Accuracy & F1 & Recall & Precision \\
 \hline
 Our & 0.832 & 0.858 &
0.778& 0.956 \\
 \hline
 Superpoint   & 0.809  & 0.833 & 0.729 & 0.978  \\
\hline
\end{tabular}
\label{table:part_count_zero}
\end{center}

\end{table}

\begin{table}
\begin{center}
\caption{Removing the unmatched parts from evaluation}
\begin{tabular}{ |c|c|c|c|c| } 
\hline
Method & Accuracy & F1 & Recall & Precision \\
 \hline
 Our & 0.918  & 0.935 & 0.945 & 0.925 \\
 \hline
 Superpoint   & 0.810  & 0.826 & 0.879 & 0.956  \\
\hline

\end{tabular}
\label{table:unmatched}
\end{center}

\end{table}

\begin{table}
\begin{center}
\caption{Basic Heuristics}
\begin{tabular}{ |c|c|c|c|c| } 
\hline
Method & Accuracy & F1 & Recall & Precision \\
 \hline
 Our & 0.89  & 0.916 & 0.907 & 0.925 \\
 \hline
 Superpoint   & 0.86  & 0.894 & 0.891 & 0.898  \\
\hline

\end{tabular}
\label{table:heuristics}
\end{center}

\end{table}




\section{Experiments and Results}

In this section, we present quantitative results on the methods present in the paper. We are using Superpoint as our baseline method. All the numbers are reported on the test dataset as mentioned in section \ref{fig:our_approach}.

Our goal is to verify  the presence of each part of the template image in the test image. During evaluation, as an input, we have a pair of test-template images and the annotation corresponding to all  parts of the template image. It's a two step process. Our first step involves the  keypoint detection and descriptors extraction from both, test and template image. We have used pretrained Superpoint as a keypoint matching algorithm in this step. To get the matched descriptors pairs we have  performed exhaustic nearest neighbour search algorithm on extracted descriptors of the images. The unmatched points and descriptors are discarded from both the images. In the second step, we have passed the matched descriptors pair through our interaction network for verification. The network verifies whether the descriptors corresponding to parts are correctly matched or not. The network predicts 1 if part is present otherwise 0.

Keypoint matching based approaches largely depend on the quality of image. If the image is of poor quality with low resolution it is very difficult for a matching algorithm to give matches. In our dataset, the quality of the test image is very poor as these are images directly taken from the shop. It is hard for keypoint matching to give matches for all parts present in the test image. We have evaluated our method in different scenarios

\begin{enumerate}
    \item \textbf{Taking unmatched part count as zero} :
    As mentioned above it's hard for a matching algorithm (in our case, Superpoint) to give matches for all parts present in the test image, so in our evaluation we take the presence value corresponding to those parts as 0. We have also calculated the Superpoint accuracy in similar conditions The results are shown in table \ref{table:part_count_zero}
    
    Keeping this evaluation metric, we perform experiments by taking the combination of different losses and various interaction operations. While our methods work best on the combination of  subtraction, multiplication, mean on the individual descriptor pairs and similarity value of two descriptors.  We also perform experiments using the other operation concat and removing the similarity score. Results are shown in table \ref{table:ablation}.
    
    \item \textbf{Removing the unmatched parts from evaluation} :
We observe a high number of false negatives in our first evaluation ref{}. As mentioned in section/ref {our method}, these high numbers of false negatives are coming from all those  parts which have zero  matches. Since these parts have zero matches and our algorithm has no control over the prediction of these parts. To check the performance of our algorithm in a more accurate way we have removed these parts from the evaluation and calculate the accuracy.  We have also calculated the Superpoint accuracy in similar conditions. The results are shown in table \ref{table:unmatched}.

    \item \textbf{Keeping basic heuristics on part count} :
We hypothesize that if a network predicts the presence of X or  more parts of the template image in the test image we can presume that all parts of the template image are present. To show that this hypothesis is true we add the heuristic on top of the final output and compare our results. We define the basic heuristics as\\

$Output\_part = 
\begin{cases}
        \text{[1]*len(Np)}, & \text{ len(Mp) $\geq$ len(Np)*P}\\
        \text{[0]*len(Np)},  & \text{otherwise}\\

\end{cases}$



\hspace{1cm} Here, Np represents the list of total parts presents in the template image, Mp is the parts with matches. P is percentage value. In our dataset, P = 0.2  gives the best number for both methods. Results are shown in table \ref{table:heuristics}

\end{enumerate}

\section{CONCLUSION}
We have established a framework for training of the verification method over local feature matches.  We have presented a method to learn the interaction between the matched descriptor pairs. Our experiments demonstrate that using the global context, local features matching can be verified correctly.  Further work will investigate the handling of noise/ wrong matches from the matching algorithm and make the verification algorithm more robust.

\end{document}